%% file: main.tex
\documentclass[journal]{IEEEtran}

\ifCLASSINFOpdf
\else
   \usepackage[dvips]{graphicx}
\fi
\usepackage{graphicx}
\usepackage{comment}
\usepackage{xcolor}
\usepackage[export]{adjustbox}
\usepackage{amsmath,amssymb,stmaryrd}  
\usepackage{pifont}
\usepackage{ulem}

\definecolor{mygray}{gray}{.9}
\begin{document}

\setlength{\abovecaptionskip}{0pt}
\setlength{\belowcaptionskip}{0pt}
\setlength{\parskip}{0pt}

\title{Enhancing Micro Gesture Recognition for Emotion Understanding via Context-aware Visual-Text Contrastive Learning}

\author{Deng Li$^{*}$, Bohao Xing$^{*}$ and Xin Liu$^\dagger$,~\IEEEmembership{Senior~Member,~IEEE}

\thanks{${*}$Equal contribution. $\dagger$Corresponding author: Xin Liu (email: xin.liu@lut.fi)}
\thanks{Bohao Xing is with the School of Electrical and Information
Engineering, Tianjin University, 300072 Tianjin, China, and also with the Computer Vision and Pattern Recognition Laboratory, Lappeenranta-Lahti University of Technology LUT, 53850 Lappeenranta, Finland (e-mail: xingbohao@tju.edu.cn).}
\thanks{Deng Li and Xin Liu are with the Computer Vision and Pattern Recognition Laboratory, Lappeenranta-Lahti University of Technology LUT, 53850 Lappeenranta, Finland (e-mail: deng.il@lut.fi, xin.liu@lut.fi).}}

\markboth{IEEE SIGNAL PROCESSING LETTERS}
{Shell \MakeLowercase{\textit{et al.}}: Bare Demo of IEEEtran.cls for IEEE Journals}
\maketitle
\begin{abstract}
Psychological studies have shown that Micro Gestures (MG) are closely linked to human emotions. MG-based emotion understanding has attracted much attention because it allows for emotion understanding through nonverbal body gestures without relying on identity information (e.g., facial and electrocardiogram data). Therefore, it is essential to recognize MG effectively for advanced emotion understanding. However, existing Micro Gesture Recognition (MGR) methods utilize only a single modality (e.g., RGB or skeleton) while overlooking crucial textual information. In this letter, we propose a simple but effective visual-text contrastive learning solution that utilizes text information for MGR. In addition, instead of using handcrafted prompts for visual-text contrastive learning, we propose a novel module called \texttt{Adaptive prompting} to generate context-aware prompts. The experimental results show that the proposed method achieves state-of-the-art performance on two public datasets. Furthermore, based on an empirical study utilizing the results of MGR for emotion understanding, we demonstrate that using the textual results of MGR significantly improves performance by 6\%+ compared to directly using video as input.
\end{abstract}


\IEEEpeerreviewmaketitle

\section{Introduction}\label{Sec:Introduction}
\input{1Introduction}

\section{Proposed method}
\input{2Proposedmethod}
\section{Experimental result}
\input{3Experimentalresult}

\section{Conclusion}
In this letter, we introduce a visual-text contrastive learning solution for Micro Gesture Recognition (MGR). Additionally, we present novel \texttt{Adaptive prompting} for context-aware prompts that integrate the contextual information of visuals. The experimental results show the superiority can be seen in Table~\ref{tab:Comparativegestures} and Table~\ref{tab:Comparativegestures_SMG}. We conducted an empirical study on how to leverage MGR results for emotion understanding. The experimental results revealed that textual predictions of MGR offer more information than other modalities, thereby improving the classification of emotions, as shown in Table~\ref{tab:Comparisonemotion}.

\bibliographystyle{IEEEtran}
\normalem
\bibliography{main.bib}
\ifCLASSOPTIONcaptionsoff
  \newpage
\fi

\end{document}

%% file: 1Introduction.tex
\IEEEPARstart{E}{motion} understanding is one of the most fundamental yet challenging tasks. It plays a crucial role in many high-level tasks such as human-computer interaction~\cite{fragopanagos2005emotion} and social robotics~\cite{tsiourti2019multimodal}. Prior research has employed different forms of data for emotion understanding such as speech~\cite{chen20183}, facial expression~\cite{mollahosseini2017affectnet,sun2020multi,yuan2023describe} and electrocardiogram (ECG) signals~\cite{chen2021clecg,mellouk2023cnn}. Although these approaches have shown promising results, they invariably use sensitive biometric data. Micro Gesture (MG) refers to the subtle body language associated with stress responses~\cite{allan1995body}. Psychological studies~\cite{aviezer2012body,de2015perception} highlight the significance of MG as a crucial clue for understanding hidden emotions. For instance, an individual may instinctively touch his or her nose when feeling anxious. In this letter, we focus on MG-based emotion understanding that,0 avoids using biometric data. MG-based emotion understanding can be divided into two tasks: 1) clip-level Micro Gesture Recognition (MGR), where the objective is to recognize the MG present in multiple clips in a video; 2) video-level emotion understanding, where the overall emotional state of the video is estimated based on the results obtained from MGR in multiple clips.

\begin{figure}[t]
    \centering
    \scriptsize
    \setlength\tabcolsep{1pt}
    \begin{tabular}{cc}
    \includegraphics[width=0.45\linewidth,cfbox=red 1pt 1pt]{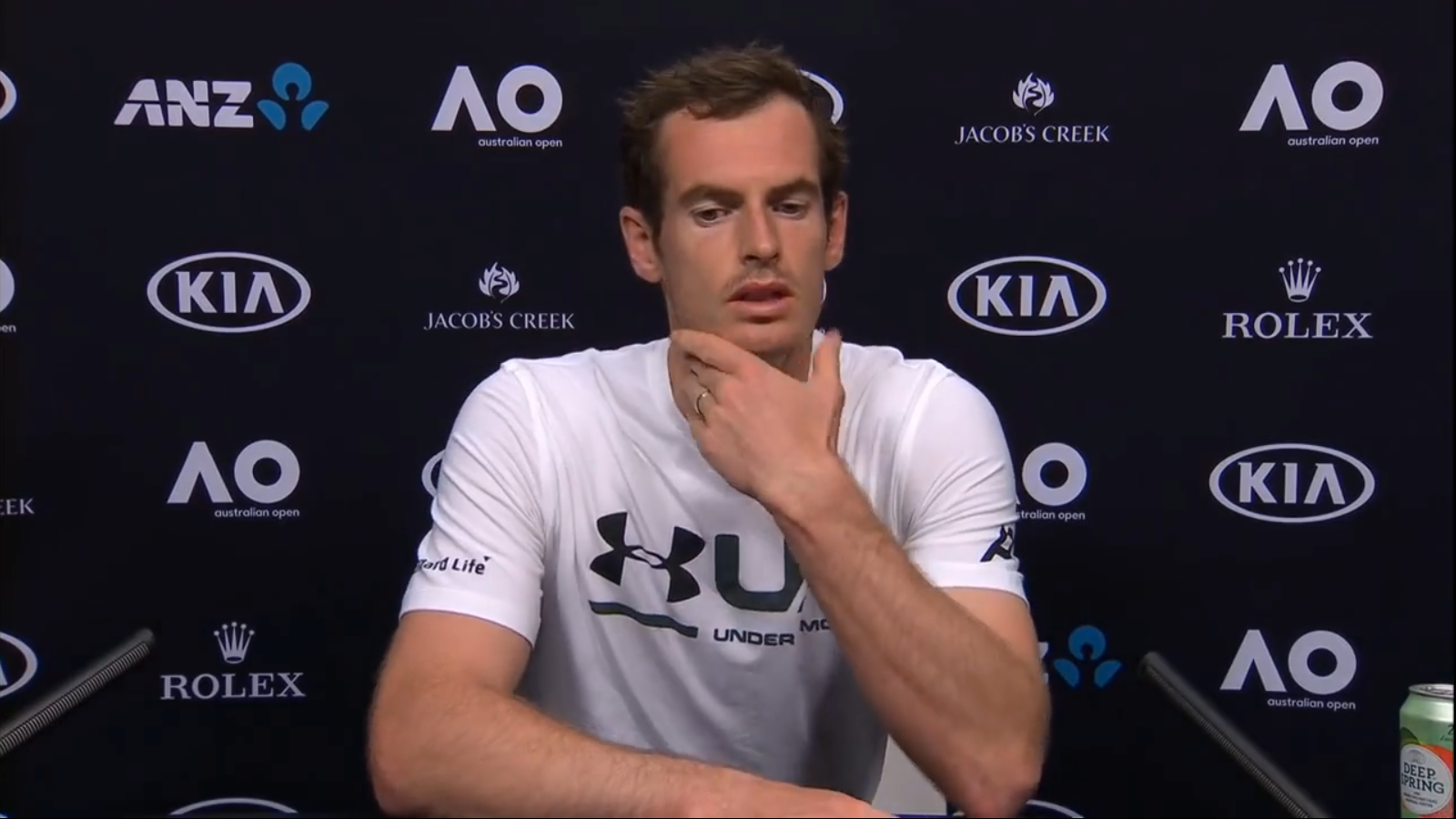}&\includegraphics[width=0.45\linewidth,cfbox=green 1pt 1pt]{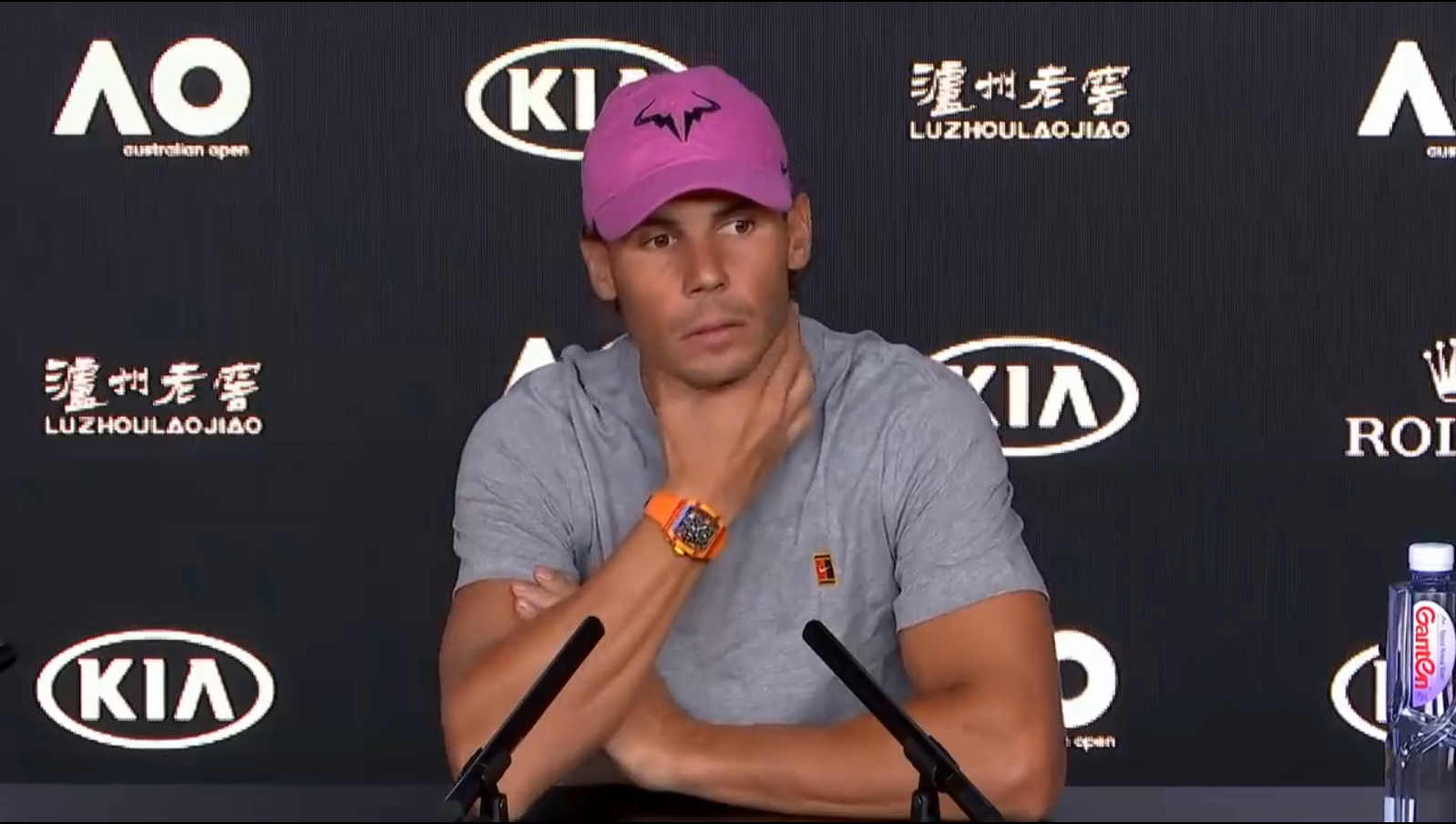}  \\
    (a) touching jaw & (b) touching or scratching neck\\
    \includegraphics[width=0.45\linewidth,cfbox=black 0.5pt 0.5pt]{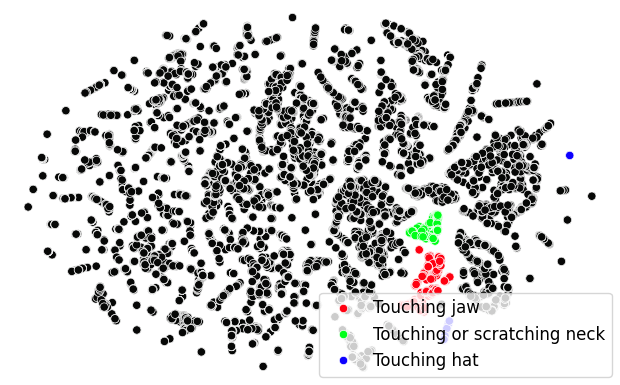}&\includegraphics[width=0.45\linewidth,cfbox=black 0.5pt 0.5pt]{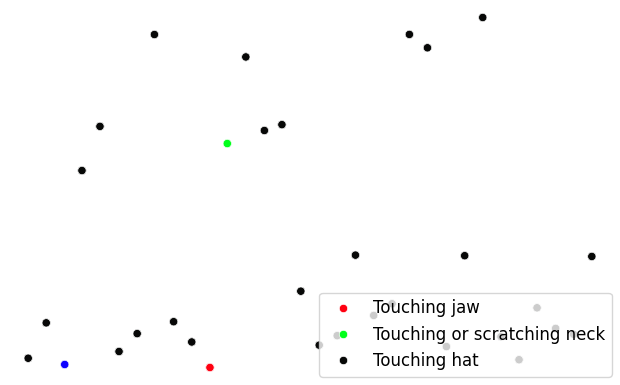}  \\
    (c) visual feature space & (d) text feature space\\
    \end{tabular}
    \caption{Some micro gestures (e.g., ``\textit{touching jaw}'' and ``\textit{touching or scratching neck}'') exhibit similarities in visual feature space, as depicted in (c), but they differ in text feature space, as illustrated in (d). Is it enough to rely solely on visual cues to recognize micro gestures? How can we leverage text information for micro gesture recognition? Best viewed digitally in color and zoomed-in.}
    \label{fig:motivation}
\end{figure}

\textbf{Clip-level micro gesture recognition.} In the era of deep learning, numerous supervised Deep Neural Networks (DNNs) methods such as RGB-based~\cite{carreira2017quo,lin2019tsm,yu2021searching} that 
skeleton-based~\cite{yan2018spatial,liu2020disentangling, peng2021spatial,liu20203d,liu20203ds} have been proposed for human action recognition. Furthermore, unsupervised learning solutions have also been explored. Liu et al.~\cite{liu2021imigue} present Sequential Variational AutoEncoder (S-VAE) for MGR. Gao et al.~\cite{gao2022cdclr} utilize contrastive learning by constructing positive/negative pairs for skeleton sequences and clips. \textcolor{black}{Shah et al.}~\cite{shah2022efficient} introduce prior knowledge from large datasets for MGR. However, these approaches rely wholly on RGB or skeleton information while overlooking textual information. Yet, textual information is crucial in MGR. We use a pre-trained \texttt{R(2+1)D} network~\cite{tran2018closer}, and \texttt{DistilBERT}~\cite{sanh2019distilbert} to extract visual and textual features of MGs, respectively. Subsequently, these features are projected into a 2D-dimensional space using t-SNE~\cite{hinton2002stochastic}, as illustrated in Fig.\ref{fig:motivation}. As one may see, ``\textit{touching jaw}'' and ``\textit{touching or scratching neck}'' are similar in the visual domain but very different in the textual domain, which prompts us to think: \textit{How can we leverage text information for micro gesture recognition?}
Recently, image-text contrastive learning methods such as CLIP~\cite{radford2021learning} have shown promising performance on many sub-stream tasks. Inspired by CLIP~\cite{radford2021learning}, we propose a simple yet effective visual-text contrastive learning solution for MGR. The concept is to pull the pairwise MG clip and corresponding MG label close to each other. To achieve this, we use a video encoder to embed the MG clip into a visual representation $\overline{v}_i$ while using a text encoder to embed the MG label into a text representation $\hat{t}_i$. Then, we optimize the model based on the similarity score. In addition, CLIP uses handcrafted prompts like ``\textit{A photo of \{object\}}'' for performing specific image-text tasks. However, handcrafted prompts fail to leverage contextual information from visuals. To solve this limitation, a novel module called \texttt{Adaptive prompting} is proposed. This module consists of Multi-Head Self-Attention (\texttt{MHSA})~\cite{vaswani2017attention} layers to capture the relationship between extracted visual and text representations. The target of \texttt{Adaptive prompting} is to generate context-aware prompts which integrate visual representations. \textcolor{black}{The comparative experiment demonstrates that the proposed method achieves State-of-The-Art (SoTA) performance in MGR (see Table~\ref{tab:Comparativegestures} and Table~\ref{tab:Comparativegestures_SMG}). }

\begin{figure}[htbp]
    \centering
    \includegraphics[width=0.55\linewidth]{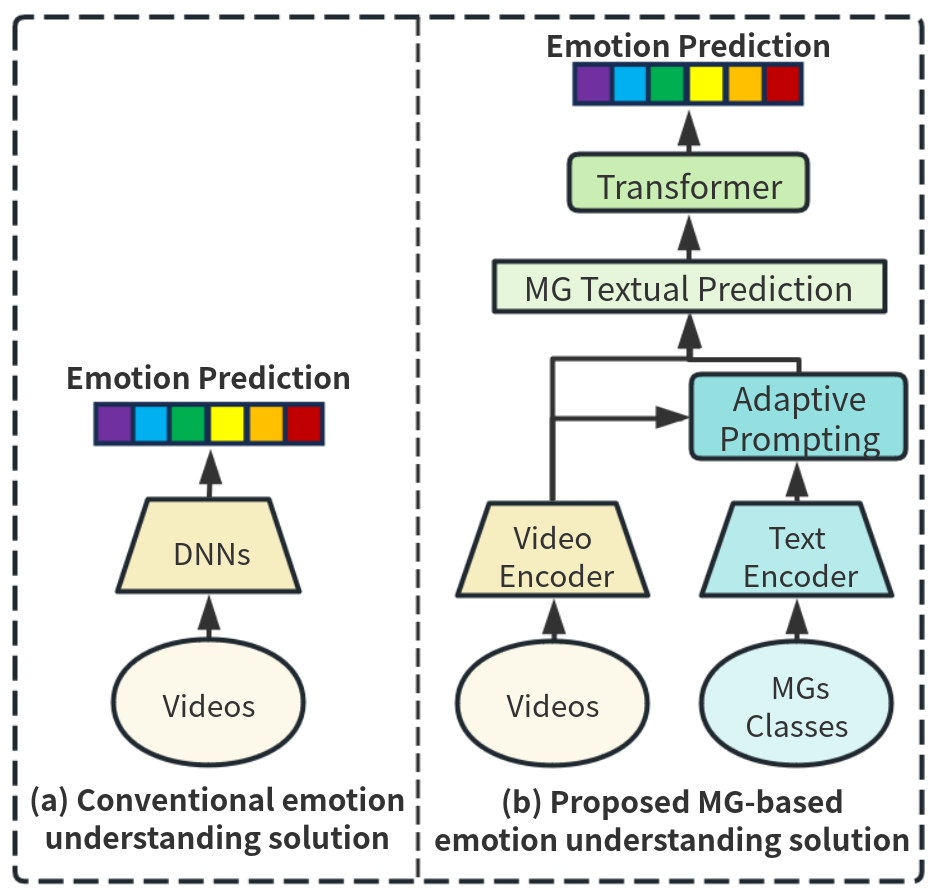}
    \caption{Comparison of different emotion understanding solutions: The proposed method initially recognizes micro gestures through visual-text contrastive learning. It subsequently utilizes the textual prediction results from micro gesture recognition for emotion understanding. }
    \label{fig:comparsion_modality_solution}
\end{figure}

\textbf{Video-level Emotion Understanding.} Conventional solutions for emotion understanding~\cite{huang2021emotion, keshari2019emotion} often rely on well-designed DNNs to extract visual features, which are then utilized to predict emotions, as illustrated in Fig.~\ref{fig:comparsion_modality_solution}. However, MG-based emotion understanding has not extensively discussed the utilization of MGR results. Therefore, we conducted an empirical study to evaluate the impact of different modalities of MGR results on emotion understanding. More precisely, we compared the following different modalities of MGR results: 1) MG visual representations, which are the extracted features of MG clips; 2) MG probability predictions, a vector of probability distributions for MG classes; 3) MG textual predictions, which represent the output of estimated MG classes. The experimental results indicate that using MG textual predictions as input surpasses other modalities in accuracy by approximately 2\% (see in Table~\ref{tab:Comparisonemotion}).

\begin{figure*}[htbp]
    \centering
    \includegraphics[width=0.7\linewidth]{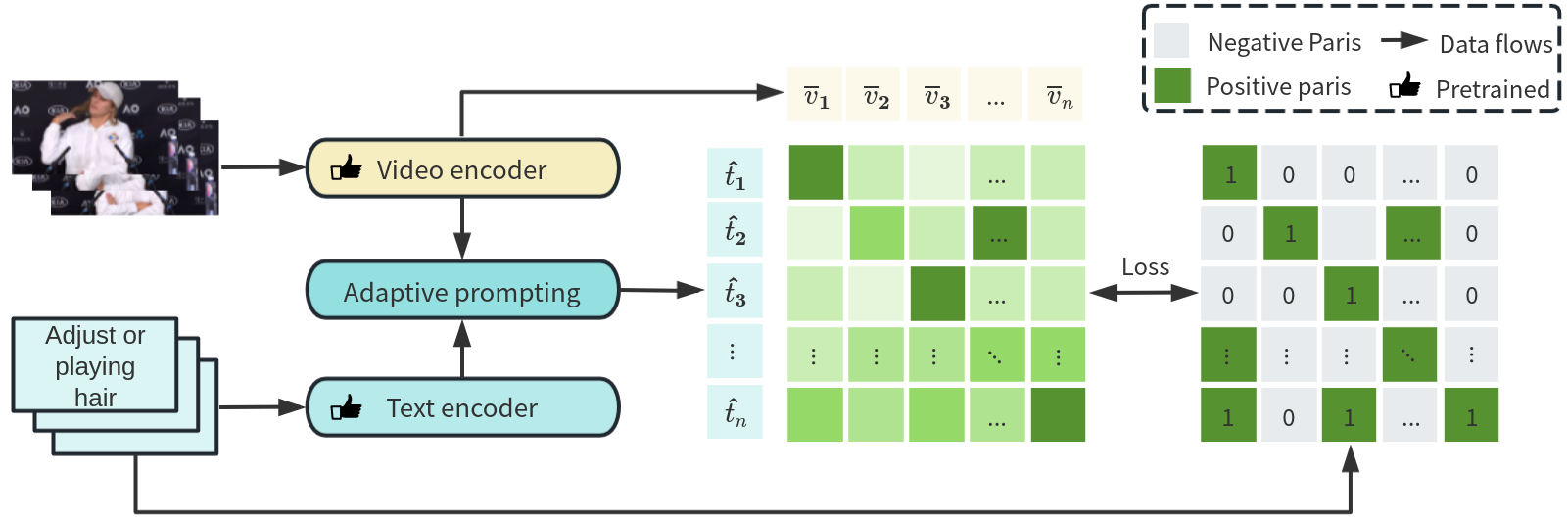}
    \caption{Structure of the clip-level micro gesture recognition. We employ a pre-trained video encoder to encode micro gesture clips as video representation $\overline{v}_i$ and a pre-trained text encoder to encode pre-defined micro gesture labels as video representation $\overline{t}_i$. The proposed \texttt{Adaptive prompting} integrates the contextual information from $\overline{v}_i$ with $\overline{t}_i$ and then generates $\hat{t}_i$. We adopt Kullback-Leibler divergence loss KL($\cdot$) to optimize the similarity score.}
    \label{fig:framework}
\end{figure*}

The contributions of this letter can be concluded as follows: 1) we propose a simple yet effective visual-text contrastive learning solution for MGR. The comparative experimental results confirm the superiority of the proposed method; 2) we design a novel module called \texttt{Adaptive prompting}, which leverages contextual information from visuals to generate context-aware prompts. An ablation study demonstrates the effectiveness of this module; 3) we discuss the different modalities of the MGR results for emotion understanding and present an empirical analysis.

%% file: 2Proposedmethod.tex
In this section, we present the proposed method for clip-level micro gesture recognition in Sec.\ref{Subsec:MG_recognition} and for video-level emotion understanding in Sec.~\ref{Subsec:Emotion_understanding}, respectively.

\subsection{Micro gesture recognition}\label{Subsec:MG_recognition}
Unlike previous single modality solutions~\cite{liu2021imigue,gao2022cdclr,shah2022efficient}, which rely solely on visual information for Micro Gesture Recognition (MGR), we propose a visual-text contrastive learning solution for MGR as shown in Fig. 3. 

\textbf{Video Encoder and Text Encoder.} The objective of the video encoder and text encoder is to embed the Micro Gesture (MG) clips and MG classes into the latent feature space. Formally, given a video $\mathbf{V}$, we divide it into several clips $V_i \in \mathbb{R}^{F \times H \times W \times 3}$, where $F$, $H$, and $W$ denote the sampled frames, height, and width, respectively. The visual representations $v_i$ are extracted by a pre-trained \texttt{R(2+1)D}~\cite{tran2018closer} network. We then use a Feedforward Neural Network (\texttt{FNN}) to project $v_i$ to $\overline{v}_i$, where $\overline{v}_i \in \mathbb{R}^{D}$ and $D$ is the projection dimension. Similar to the video encoder, we employ the pre-trained \texttt{DistillBert}~\cite{sanh2019distilbert} and \texttt{FNN} to obtain the text representation $\overline{t}_i \in \mathbb{R}^{D}$. The value of $D$ is set to 256 throughout the paper.

\textbf{Adaptive Prompting.} 
CLIP employs handcrafted prompts (e.g., \textit{a photo of \{label\}}) for sub-tasks such as image classification. However, this handcrafted prompting approach overlooks contextual cues. Therefore, we propose a module called \texttt{Adaptive prompting} to address this limitation by utilizing Multi-Head Self-Attention (\texttt{MHSA})~\cite{vaswani2017attention}. In \texttt{MHSA}, each token in a feature attends to all other tokens to compute a representation for itself, allowing for the capture of internal feature correlations. Thus, we use the text representation $\overline{t}$ to derive $Q_i = \overline{t}_i W_{i}^{q}$, and the visual representation $\overline{v}_i$ to derive $K_i = \overline{v}_i W_{i}^{k}$ and $V_i = \overline{v}_i W_{i}^{v}$, as shown below:
\begin{equation}
    {\rm att} = \overline{t}_i + {\rm softmax}(\frac{Q_i K_{i}^T}{\sqrt{d_k}})V_{i}
\end{equation}
where $W_{i}^{q}$, $W_{i}^{k}$, and $W_{i}^{v}$ represent these weight matrices. Next, we enhance the text presentation $\hat{t}_i$ by:
\begin{equation}
    \hat{t}_i = \overline{t}_i + \lambda({\rm att} + {\rm FNN}({\rm att}))
\end{equation}
where $\lambda$ is a learnable parameter. The proposed \texttt{Adaptive prompting} allows the text representation to integrate prompts that contain the visual contextual information from the visual representation. An ablation study of \texttt{Adaptive prompting} is presented in Table~\ref{tab:Ablation_study}.

\textbf{Alignment.} CLIP~\cite{radford2021learning} optimizes similarity scores by cross-entropy loss. However, we argue that it may not be suitable. Because multiple clips may correspond to the same MG within a batch, MGR is not a 1-in-1 classification problem. Thus, we decided to use Kullback-Leibler divergence loss ${\rm KL}(\cdot)$ to measure the difference between the probability distributions of the predicted similarity score and target similarity score:
\begin{equation}
    Loss = \frac{1}{2}({\rm KL}(S^{visual},GT) + {\rm KL}(S^{text},GT))
\end{equation}
where $S^{visual} = (\overline{v} \cdot \hat{t}^{T})/{\tau}, S^{text} = (\hat{t} \cdot \overline{v}^{T})/{\tau}$ and $GT \in \mathbb{R}^{n \times n}$ is defined as below:
\begin{equation}
    GT[i,j] = 
    \begin{cases}
        1, & \text{if } T_j = T_i \\
        0, & \text{otherwise}
    \end{cases}
\end{equation}
where $\mathbf{T} = [T_1, T_2, ..., T_n]$ is the MG labels in batch. After aligning the video encoder with the text encoder, we use the aligned video encoder as a frozen intermediate network to finetune a Multi-Layer Perceptron (\texttt{MLP}) classifier for MGR.

\subsection{Emotion understanding}\label{Subsec:Emotion_understanding}
After implementing MGR, we use the textual predictions of clips $V_i$ and employ the \texttt{Transformer}~\cite{vaswani2017attention} model to estimate the emotion. Formally, the estimated MG $T^{*} = \{t_{1}^{*}, t_{2}^{*}, ..., t_{n}^{*}\}$, where $t_{i}^{*}$ is the prediction for clips $V_i$. The estimations are embedded by an embedding layer as $\{E_{t1}^{*}, E_{t2}^{*}, ..., E_{tn}^{*}\}$. We prepend the trainable [class] token $t_{\text{class}}^{*}$. The input is denoted as:
\begin{equation}
    \overline{T}^{*} = [t_{\text{class}}^{*}, E_{t1}^{*}, E_{t2}^{*}, ..., E_{tn}^{*}] + e^{\text{pos}}
\end{equation}
where $e^{\text{pos}}$ denotes positional encoding. Then, we feed $\overline{T}^{*}$ into the \texttt{Transformer} with the following \texttt{FNN} to predict the emotion. Please refer to the supplementary material for more details of the architecture of \texttt{Transformer}.

%% file: 3Experimentalresult.tex
\subsection{Experimental setting}
\textcolor{black}{The training of the proposed method consists of three stages: 1) we conduct joint training of the video encoder and text encoder. Each clip has 32 frames obtained by uniform sampling, and the spatial resolution is $224 \times 224$. The learning rates for the video and text encoders are $1e-4$ and $1e-5$, respectively. The temperature parameter $\tau$ is set to 0.05; 2) we finetune a two-layer \texttt{MLP} classifier for MGR by using cross-entropy loss with learning rate $1e-3$;} 3) we train the \texttt{Transformer} to estimate emotion from textual micro gesture predictions, employing cross-entropy loss with a learning rate of $1e-5$. Throughout the paper, we consistently apply the \texttt{AdamW}~\cite{loshchilov2017decoupled} optimizer. The implementation is available at https://github.com/linuxsino/Visual-Text-MG.

To evaluate the effectiveness of the proposed model, we selected the following datasets: 1) iMiGUE~\cite{liu2021imigue} collected MG samples of different tennis players when they participated in post-match interviews. There are 18,499 wild-collected MG samples for MGR and 359 videos for emotion understanding. It is noted that the start time and end time of clips in each video are predefined in the provided protocol for video-level emotional understanding; 2) SMG~\cite{chen2023smg} is the new benchmark dataset for MGR, collected from 40 subjects who narrated both fake and real stories. This dataset consists of 3,692 lab-collected MG samples. Following the methodology outlined in~\cite{liu2021imigue}, we employed accuracy top-1 and top-5 for MGR and accuracy top-1 for emotion understanding.

\subsection{Ablation study}
Before comparing the proposed method with State-of-The-Art (SoTA) approaches, there are three important questions that need to be addressed. We conducted experiments to investigate these questions using the dataset iMiGUE~\cite{liu2021imigue}. We adopt the~\texttt{R(2+1)D}~\cite{tran2018closer} network and RGB input as the baseline for the ablation experiment, as shown in Table~\ref{tab:Ablation_study}.

\textbf{Does visual-text contrastive learning help?}
To leverage textual information for MGR, we propose a visual-text contrastive learning solution. As shown in Table~\ref{tab:Ablation_study}, the baseline model exhibits suboptimal performance. Notably, even using the same video encoder, the proposed visual-text contrastive learning solution significantly enhances the performance, increasing the accuracy top-1 from 38.17\% to 64.60\%. The resulting confusion matrix of baseline (w/o text information) and ours (with text information) are presented in Fig.~\ref{fig:confusion matrix}.

\textbf{Does adaptive prompting help?} 
We introduce the module \texttt{Adaptive prompting} to generate context-aware prompts instead of using the handcrafted prompts in CLIP~\cite{radford2021learning}. To evaluate the effectiveness of the proposed module \texttt{Adaptive prompting}, we designed three handcrafted prompting templates, namely, ``\textit{a photo of \{label\}}'', ``\textit{a video of \{label\}}'' and ``\textit{an action of \{label\}}'' for comparison. As shown in Table~\ref{tab:Ablation_study}, the utilization of the module \texttt{Adaptive prompting} resulted in performance improvement (1\%+ in top-1 accuracy) when compared to the handcrafted prompting.

\begin{table}[htbp]
    \centering
    \footnotesize
    \setlength\tabcolsep{1pt}
    \caption{\textcolor{black}{Ablation study of proposed micro gesture recognition solution on iMiGUE dataset}}
    \begin{tabular}{l|rrr}\hline\hline
         &\textbf{Acc. Top-1}$(\%)\uparrow$&\textbf{Acc. Top-5}$(\%)\uparrow$ \\\hline
         Baseline&38.17&72.99\\
         + Visual-text contrastive learning&64.60&85.26\\
         + Handcrafted prompting&65.01&85.31\\
         + Adaptive prompting&66.06&85.61\\
         + Finetune&\textbf{66.12}&\textbf{87.27}\\\hline\hline
         \multicolumn{3}{l}{Note: No handcrafted prompts are used in + Adaptive prompting.}\\
    \end{tabular}
    \label{tab:Ablation_study}
\end{table}
\begin{figure}[htbp]
    \centering
    \setlength\tabcolsep{1pt}
    \begin{tabular}{cc}
         \includegraphics[width=0.298\linewidth,cfbox=black 0.5pt 0.5pt]{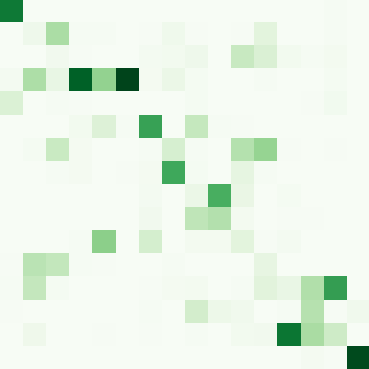}&\includegraphics[width=0.35\linewidth,cfbox=black 0.5pt 0.5pt]{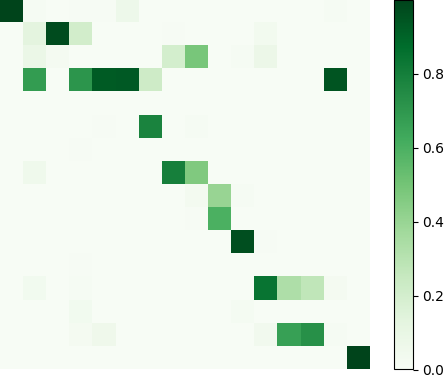} \\
         (a) w/o text info. & (b) w/ text info.
    \end{tabular}
    \caption{\textcolor{black}{Comparison of the resulting confusion matrix for MGR. 'W/o text info.' denotes the baseline model, and 'w/ text info.' denotes the proposed visual-text model. The X-axis represents the model's predicted class, while the Y-axis represents the actual class.}}
    \label{fig:confusion matrix}
\end{figure}

\textbf{Does textual prediction of MGR work better?} Various modalities of the results of MGR can be used for emotion understanding: 1) visual representations, which is the output of the video encoder; 2) the probability vector of MG; 3) the textual prediction of MG. Therefore, we train the \texttt{Transformer} with different inputs and report the performance in Table ~\ref{tab:Comparison_modality_emotion}. As one may see, using MG textual prediction is superior to the other modalities. This result indicates that the textual features contain more information than the probability vectors and visual representations for emotion classification.

\begin{table}[htbp]
    \centering
    \footnotesize
    \setlength{\tabcolsep}{3pt}
    \caption{Comparison of different modalities of MGR for emotion understanding on iMiGUE dataset}
    \begin{tabular}{l|rrr}
    \hline\hline
         Modality&\textbf{Accuracy Top-1}$(\%)\uparrow$ \\\hline
         
         MG visual representations&59.11\\
         MG probability vector&61.45 \\
         MG textual prediction&\textbf{63.29} \\
    \hline\hline
    \end{tabular}
    \label{tab:Comparison_modality_emotion}
\end{table}

\subsection{Comparative study}
\textbf{Clip-level micro gestures recognition.}
This experiment focuses on recent DNN solutions with different learning strategies and input modalities. We followed the training and testing protocol of iMiGUE~\cite{liu2021imigue} and SMG~\cite{chen2023smg,chen2019analyze}. The experimental results are presented in Table~\ref{tab:Comparativegestures} and Table~\ref{tab:Comparativegestures_SMG}. It is clear that the proposed method demonstrates SoTA performance on two datasets. For instance, in the iMiGUE dataset, our method significantly outperforms TSM~\cite{lin2019tsm} by a substantial margin (5\%+ in top-1 accuracy). This result demonstrates the effectiveness of the proposed visual-text contrastive learning solution.

\begin{table}[htbp]
    \centering
    \footnotesize
    \setlength{\tabcolsep}{3pt}
    \caption{\textcolor{black}{Comparison of the proposed method with the SoTA on the iMiGUE dataset}}
    \begin{tabular}{l|rrr}
    \hline\hline
         \textbf{Method}&\textbf{Modality}&\textbf{Acc. Top-1}$(\%)\uparrow$&\textbf{Acc. Top-5}$(\%)\uparrow$ \\\hline
         P\&C~\cite{su2020predict}&Pose&31.67&64.93\\
         U-S-VAE~\cite{liu2021imigue}&Pose&32.43&64.30\\          
         EDGCN~\cite{shah2022efficient}&Pose&37.50&-\\        
         CdCLR~\cite{gao2022cdclr}&Pose&39.38&76.40\\
         ST-GCN~\cite{yan2018spatial}&Pose&46.97&84.09\\
         MS-G3D~\cite{liu2020disentangling}&Pose&54.91&89.98\\      
         MA-GCN~\cite{yuan2023mstcn}&Pose&55.32&89.92\\
         CE-CN~\cite{shah2023representation}&Pose&56.12&90.01\\
         MSTCN-VAE~\cite{yuan2023mstcn}&RGB&47.69&56.36\\
         I3D~\cite{carreira2017quo}&RGB&34.96&63.69\\
         TSM~\cite{lin2019tsm}&RGB&61.10&\textbf{91.24}\\
         Ours&RGB + Text&\textbf{66.12}&87.27\\
         \hline\hline
    \end{tabular}
    \label{tab:Comparativegestures}
\end{table}

\begin{table}[htpb]
    \centering
    \footnotesize
    \setlength{\tabcolsep}{3pt}
    \caption{\textcolor{black}{Comparison of the proposed method with the SoTA on the SMG dataset}}
    \begin{tabular}{l|rrr}
        \hline\hline \textbf{Method}&\textbf{Modality}&\textbf{Acc. Top-1}$(\%)\uparrow$&\textbf{Acc. Top-5}$(\%)\uparrow$ \\\hline
         ST-GCN~\cite{yan2018spatial}&Pose&41.48&86.07\\
         EDGCN~\cite{shah2022efficient}&Pose&47.90&-\\
         SHift-GCN~\cite{cheng2020skeleton}&Pose&55.31&87.34\\
         TRN~\cite{xu2019temporal}&Pose&59.51&88.53\\
         MS-G3D~\cite{liu2020disentangling}&Pose&64.75&91.48\\
         I3D~\cite{carreira2017quo}&RGB&35.08 &85.90\\
         MSTCN-VAE~\cite{yuan2023mstcn}&RGB&42.59&49.54\\
         TSM~\cite{lin2019tsm}&RGB&58.69&83.93\\
         Ours&RGB + Text&\textbf{65.08}&\textbf{91.64}\\
         \hline\hline
    \end{tabular}
    \label{tab:Comparativegestures_SMG}
\end{table}

\textbf{Video-level emotion understanding.}
We conducted an experiment on video-level emotion understanding using iMiGUE [17]. TSM~\cite{lin2019tsm}, I3D~\cite{carreira2017quo}, GCN~\cite{yan2018spatial}, and MS-G3D~\cite{liu2020disentangling} employed the same configurations as the clip-level MGR experiment, but with the modification of a change from clip to video input. For MG-based solutions, we chose unsupervised U-S-VAE~\cite{lin2019tsm} and supervised TSM~\cite{lin2019tsm}, which works well on MGR as the baseline. Then, the results of various modalities of clip-level MGR were fed into \texttt{LSTM} or the \texttt{Transformer}. The results of the experiment are presented in Table~\ref{tab:Comparisonemotion}. The results clearly indicate the superior performance of MG-based solutions compared to RGB-based and Pose-based solutions. Our inference is that directly utilizing RGB/Pose features proves challenging in extracting pertinent information for emotion understanding. Furthermore, the results affirm that textual predictions outperform other MG modalities by approximately 2\% in Top-1 accuracy.

\begin{table}[htbp]
    \centering
    \footnotesize
    \setlength{\tabcolsep}{3pt}
    \caption{Comparison of the proposed method with SoTA in emotion understanding on iMiGUE dataset}
    \begin{tabular}{l|rrr}
    \hline\hline
    \textbf{Method}&\textbf{Modality}&\textbf{Accuracy Top-1}$(\%)\uparrow$  \\\hline
         $\dagger$TSM~\cite{lin2019tsm}&RGB&53.00\\
         $\dagger$I3D~\cite{carreira2017quo}&RGB&57.00\\
         $\dagger$ST-GCN~\cite{yan2018spatial}&Pose&50.00\\
         $\dagger$MS-G3D~\cite{liu2020disentangling}&Pose&55.00\\
         $\dagger$U-S-VAE~\cite{liu2021imigue} + LSTM&MG probability vector&55.00\\
         TSM~\cite{lin2019tsm} + LSTM&MG probability vector&60.37\\
         TSM~\cite{lin2019tsm} + Transformer&MG probability vector&61.41\\
         Ours&MG textual prediction&\textbf{63.29}\\
         \hline\hline
         \multicolumn{3}{l}{Note: $\dagger$ means the result is from \cite{liu2021imigue}.}\\
    \end{tabular}
    \label{tab:Comparisonemotion}
\end{table}